\documentclass[9pt]{extarticle}  
\usepackage{spconf,amsmath,graphicx,hyperref}
\usepackage{amssymb}
\usepackage{booktabs}
\usepackage{multirow}
\usepackage{hyperref}
\usepackage{cite}
\usepackage{makecell}

\title{EMPATHY OMNI: ENABLING EMPATHETIC SPEECH RESPONSE GENERATION THROUGH LARGE LANGUAGE MODELS
}
\name{Haoyu Wang$^{\star \dagger}$ \quad Guangyan Zhang$^{\dagger}$ \quad Jiale Chen$^{\star}$ \quad Jingyu Li$^{\dagger}$ \quad Yuehai Wang$^{\star}$ \quad Yiwen Guo$^{\ddagger}$}
  
\address{$^{\star}$ Zhejiang University \\
    $^{\dagger}$ LIGHTSPEED \\
    $^{\ddagger}$ Independent Researcher}

\begin{document}
\maketitle
\begin{abstract}
With the development of speech large language models (speech LLMs), users can now interact directly with assistants via speech. However, most existing models only convert response content into speech without fully capturing the rich emotional cues in user queries, where the same sentence may convey different meanings depending on the expression. Emotional understanding is thus essential for improving human–machine interaction.   
Most empathetic speech LLMs rely on massive datasets, demanding high computational cost. A key challenge is to build models that generate empathetic responses with limited data and without large-scale training.  
To this end, we propose Emotion Omni, a model that understands emotional content in user speech and generates empathetic responses. We further developed a data pipeline to construct a 200k emotional dialogue dataset supporting empathetic speech assistants.  
Experiments show that Emotion Omni achieves comparable instruction-following ability without large-scale pretraining, while surpassing existing models in speech quality (UTMOS:4.41) and empathy (Emotion GPT Score: 3.97). These results confirm its improvements in both speech fidelity and emotional expressiveness.  

Demos are available at \texttt{\url{https://w311411.github.io/omni_demo/}}.

\end{abstract}

\section{Introduction}
\label{sec:intro}

With the emergence of systems such as GPT-4o~\cite{hurst2024gpt,cui2024recent}, integrating speech capabilities with large language models (LLMs) has substantially improved user experience and broadened application scenarios, including customer service, education, personal assistance, and therapeutic support. Recent end-to-end multimodal voice interaction models, such as Mini-Omni~\cite{miniomni}, LLaMA-Omni~\cite{llamaomni}, and VocalNet~\cite{wang2025vocalnet} have moved beyond the conventional pipeline of separate speech-to-text transcription followed by text generation and text-to-speech. By directly generating textual and speech responses from voice commands, these models streamline interaction, reduce latency, and achieve competitive question-answering performance while improving speech quality and emotional alignment with substantially less training data.

Beyond semantic question answering, effective speech interaction requires robust modeling of paralinguistic information ~\cite{an2024funaudiollm,huang2024audiogpt,ding2025kimi}. Paralinguistic cues—including the speaker’s emotions and intentions—are central to everyday communication and can fundamentally alter the meaning of the same utterance when delivered with different affect. Without accurately interpreting a user’s emotional intent and responding in an empathetic manner, a speech system may misread the situation or produce inappropriate outputs, degrading user trust and experience. This limitation is especially consequential in settings that demand high emotional intelligence, such as mental health support, customer service, and educational tutoring, where empathetic communication is crucial for rapport and effectiveness. While prosodic and emotional dialogue systems have advanced ~\cite{styletalker,liu2024generative,wang-etal-2024-blsp}, and recent speech LLMs have begun to address empathetic response generation ~\cite{zeng2024glm,chen2024slam, wang2025opens2s}, but these approaches depend on implicitly learning emotional consistency from datasets without explicit emotional modeling, requiring high-quality emotional dialogue data and extensive training. Furthermore, the limited availability of emotional dialogue data and prohibitive annotation costs hinder the development of empathetic capabilities.

\begin{figure*}[t]
    \centering
    \includegraphics[width=1.0\textwidth]{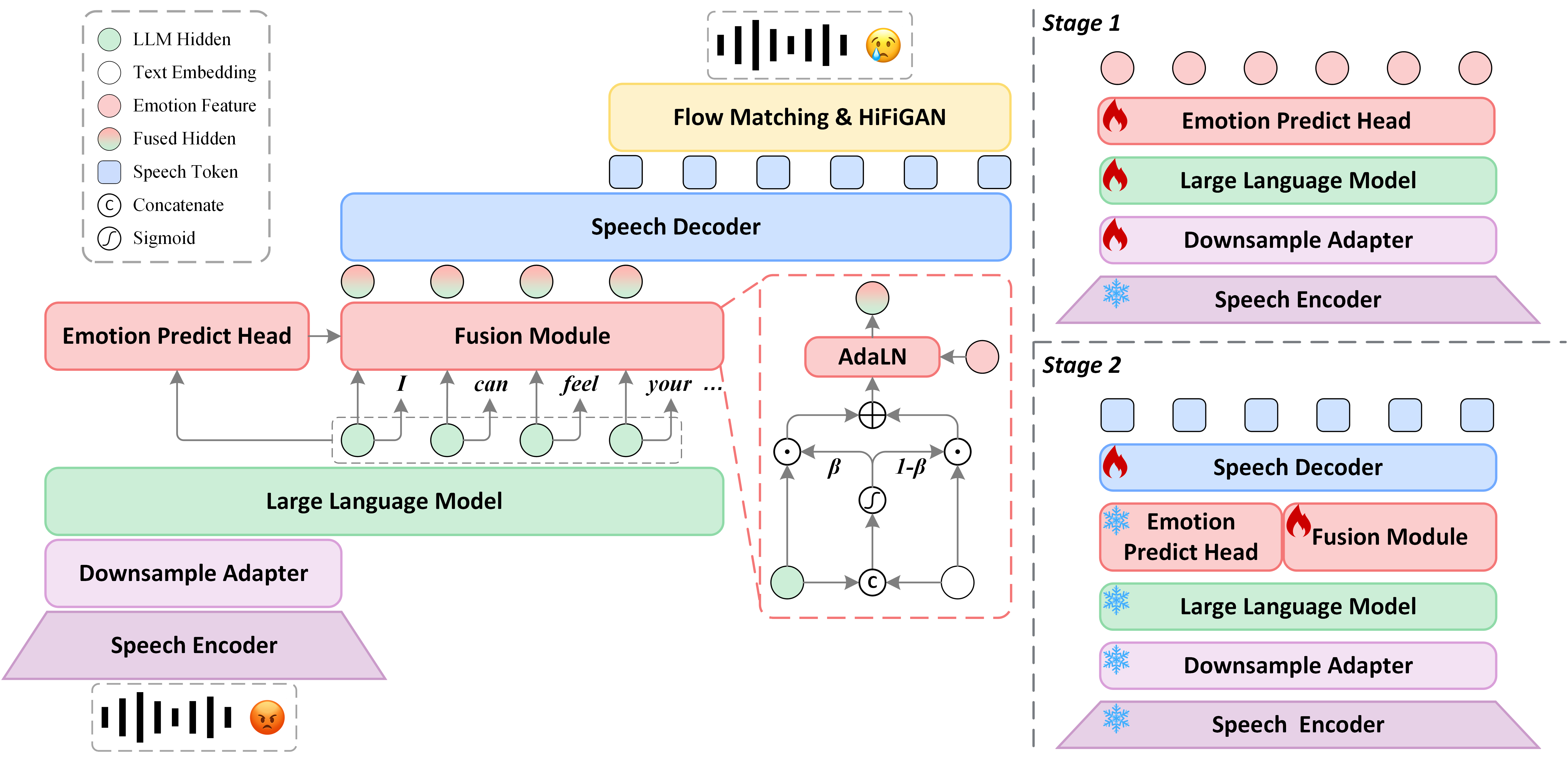}  
    \caption{Model Architecture and Training Process. In Stage 1, we fine-tune the LLM backbone with LoRA to align speech and text, focusing on understanding speech cues and generating empathetic responses. In Stage~2, the text embeddings are first fed into the speech decoder to train basic speech generation. Subsequently, the fusion module combines the LLM hidden states, text embeddings, and emotional cues to achieve better emotion control and improved speech quality.
}  
    \label{fig:model}
\end{figure*}

To address these challenges, we propose Emotion Omni, an end-to-end model that explicitly leverages emotional features in user speech to recognize affect and control the generation of empathetic responses. By jointly conditioning on semantic and emotional representations from the input speech, the model generates both the textual response and the emotional features needed to control the tone and emotion of the synthesized speech. This design separates the modeling of semantics and emotions to avoid feature entanglement. By predicting the emotion trajectory of the response, it achieves fine-grained control over speech emotion, enabling natural transitions and variations within the same utterance and ensuring more accurate and consistent emotional expression in synthesis. Complementing the model, We develop a cost-effective, scalable emotional dialogue generation pipeline using open-source TTS. The pipeline employs GPT-4o and CosyVoice2 to create emotionally consistent query-response pairs across domains, building a multi-speaker dataset filtered by ASR-WER and emotion recognition for quality assurance. Additionally, we augment synthetic data with real emotional dialogues to enhance diversity and generalization. Real dialogues provide natural emotional patterns while synthetic data expands scale and domain coverage. This yields a 200k emotional dialogue dataset enabling empathetic speech assistant development without extensive manual annotation. 
Our contributions are twofold. We present an emotion-driven speech LLM that jointly recognizes users’ emotional intent and generates empathetic speech responses, enhancing emotional understanding and response capability in an end-to-end setting. We also construct an emotional dialogue dataset that combines real and synthetic data through a scalable pipeline, significantly reducing development costs and manual annotation requirements while leveraging real data with natural emotional patterns to improve generalization performance. Through these innovations, we build a low-cost, efficient, and reproducible empathetic speech LLM, reducing the required training data and computational resources and providing a flexible solution for future empathetic speech interaction systems.

    

Through these innovations, we have successfully built a low-cost, efficient, and reproducible empathetic speech LLM, significantly reducing the required training data and computational resources, and providing a flexible and efficient solution for future empathetic speech interaction systems.

\section{Methods}
We adopt a two-tower end-to-end architecture similar to prior omni-directional speech LLMs~\cite{llamaomni}, in which a LLM performs speech understanding and text response generation, and a causal speech decoder produces acoustic tokens conditioned on linguistic and paralinguistic cues. To explicitly model paralinguistic information, we extract complementary content and emotion features with two pretrained speech encoders. After temporal downsampling, these features are fused and projected to the LLM hidden space to condition understanding and generation. The LLM outputs both response tokens and a token-synchronous emotion trajectory that serves as an explicit affective plan. The speech decoder then conditions on the response token embeddings, the corresponding LLM hidden states, and the predicted emotion trajectory to autoregressively generate speech tokens. \autoref{fig:model} outlines the architecture and training flow. Training proceeds in two stages. In Stage 1, we train the downsampling adapters and fine-tune the LLM backbone with LoRA\cite{hu2022lora} to align speech and text for empathetic response generation and to learn token-synchronous emotion planning. In Stage 2, we train the speech decoder and enable gated fusion of LLM hidden states and text embeddings together with adaptive emotional conditioning to improve naturalness and expressiveness.

\subsection{Speech Encoder}
\label{ssec:Speech Encoder}

Given a speech query, we run two pretrained encoders in parallel to extract complementary features: a semantic encoder for linguistic content and an emotion encoder for paralinguistic cues. Because the native frame rates are high, we attach lightweight downsampling adapters implemented with convolutional layers to reduce the frame rate to 10 Hz, which improves efficiency while retaining salient prosodic dynamics. To preserve the pretrained capacities and stabilize optimization, both encoders are kept frozen. We train only the downsampling adapters and subsequent projection layers. At each frame, we concatenate the downsampled semantic and emotion vectors and project the result to the LLM hidden size, yielding a synchronized sequence of speech embeddings that jointly represent what is spoken and how it is expressed. These embeddings condition the LLM during understanding and response generation.

\subsection{Large Language Model}



We employ a pretrained instruction-following LLM as the backbone due to its strong generation capabilities. 
To enable explicit affective control, we augment the LLM with an emotion prediction head that maps the last-layer hidden state of each generated token to an emotion feature vector, thereby producing a token-synchronous emotion trajectory for the response. 

During training, we derive ground-truth emotion trajectories $\mathcal{H}_{emo}$ from the target response waveform using the same emotion encoder employed on the input branch. 
We align the frame-level emotion features $\mathcal{H}_{emo}$ with the output token sequence using Dynamic Time Warping (DTW)\cite{sakoe2003dynamic}. 
Formally, the cumulative alignment cost is computed as:
\begin{align}
D(i,j) &= d(\mathcal{H}_{emo}[i], \, t_j) \nonumber \\
&\quad + \min \{ D(i-1,j), \, D(i,j-1), \, D(i-1,j-1) \}
\end{align}

where $d(\cdot,\cdot)$ is the distance metric (e.g., squared Euclidean), 
$t_j$ denotes the $j$-th token position, and $D(i,j)$ is the minimal alignment cost between the first $i$ frames and the first $j$ tokens. 
The aligned ground-truth emotion vector for token $j$ is then obtained from the DTW warping path $\pi$:
\begin{equation}
\mathcal{H}^{target}_{emo}[j] = \frac{1}{|\pi(j)|} \sum_{i \in \pi(j)} \mathcal{H}_{emo}[i],
\end{equation}

where $\pi(j)$ is the set of frame indices aligned to token $j$ by DTW. 
The training objective is multi-task: we minimize the next-token cross-entropy loss for language modeling and an auxiliary regression loss on the aligned emotion vectors that combines mean squared error with a cosine similarity term. 
A scalar weight balances the auxiliary loss against the language modeling loss. 
This encourages the LLM to produce responses that are semantically appropriate and paired with an explicit, temporally aligned emotion plan. 
At inference, the LLM emits text tokens and their corresponding emotion vectors without additional predictors or look-ahead.

\subsection{Speech Decoder}

The speech decoder comprises a projection layer followed by a causal Transformer with six decoder blocks. It autoregressively predicts discrete speech tokens under joint textual and emotional conditioning. At each decoding step, the decoder receives two linguistic signals aligned to the current position: the embedding of the current response token and the corresponding LLM hidden state. The token embedding provides fine-grained lexical content, while the hidden state supplies broader semantic and discourse context. We integrate them with a gating fusion mechanism that computes a sigmoid gate from their concatenation and forms a convex combination, allowing the model to balance content and context at each step dynamically. During inference, the LLM hidden state is taken from the autoregressively generated last hidden state, and the response token is sampled from this state; during training, the token is the ground truth, and the hidden state is obtained by teacher forcing.



Emotional control is provided by the token-synchronous emotion trajectory predicted by the LLM. The decoder conditions on these vectors via adaptive layer normalization (AdaLN)\cite{perez2018film}, which generates per-step scale and shift parameters from the aligned emotion features and applies them to the normalized fused textual representation. This modulation injects affective control in a stable, factorized manner, preserving linguistic content while enabling fine-grained prosody control. During training, we first condition on ground-truth emotion features extracted from the target speech to provide clean supervision, and then progressively mix in the predicted trajectory to mitigate train–test mismatch. The decoder is trained to predict the next speech token with a standard cross-entropy loss over the tokenizer vocabulary.

The entire fusion process is formulated as follows:
\begin{equation}
\mathcal{H}_{\text{adapted}} = \text{AdaLN}\big(\text{GateFusion}(\mathcal{H}_{\text{text}}, \mathcal{H}_{\text{hidden}}), \hat{\mathcal{H}}_{\text{emo}}\big)
\end{equation}
where $\hat{\mathcal{H}}_{\text{emo}}$ denotes the predicted emotional trajectory.



The speech tokens produced by the decoder are converted into mel-spectrograms using a chunk-aware causal flow-matching model~\cite{lipman2022flow}, which supports streaming synthesis. Specifically, every $W$ generated speech tokens are grouped into a chunk for spectrogram generation to balance latency and quality. The resulting mel-spectrograms are then transformed into waveforms by a HiFi-GAN vocoder~\cite{kong2020hifi}. Both the flow-matching model and the vocoder are adopted from pretrained CosyVoice2 for efficient and high-fidelity synthesis.

\begin{table*}[htbp]
\centering
\scalebox{1}{
\begin{tabular}{l|ccccc|c}
\hline
\multirow{2}{*}{\textbf{Model}} & \multicolumn{5}{c|}{\textbf{VoiceBench}} & \textbf{Speech Quality} \\ \cline{2-6} \cline{7-7}
 & \textbf{Alpaca $\uparrow$} & \textbf{Common $\uparrow$} & \textbf{IFEval $\uparrow$} & \textbf{WildVoice $\uparrow$} & \textbf{SD-QA $\uparrow$} & \textbf{UTMOS $\uparrow$} \\ \hline
Qwen2-Audio\cite{chu2023qwen} & 3.74 & 3.43 & 26.33 & 3.01 & 35.71 & - \\ 
Moshi\cite{defossez2024moshi} & 2.01 & 1.60 & 10.12 & 1.30 & 15.64 & 3.81 \\ 
GLM-4-Voice\cite{zeng2024glm}& \textbf{3.97} & 3.42 & 25.92 & \textbf{3.18} & 25.92 & 3.48 \\
LLaMA-Omni\cite{llamaomni} & 3.70 & 3.46 & 14.87 & 2.92 & \textbf{39.69} & 3.98 \\ \hline
Ours & 3.84 & \textbf{3.47} & \textbf{27.89} & 3.13 & 36.87 & \textbf{4.41} \\ \hline
\end{tabular}
}
\caption{Performance comparison on VoiceBench and speech quality metrics}
\label{tab:performance_comparison}
\end{table*}

\begin{table*}[!t]
\centering
\scalebox{1.1}{
\begin{tabular}{c|c|c|c}
\hline
Test Set & \textbf{Emotion GPT Score} $\uparrow$ & \textbf{Speech Emotion MOS} $\uparrow$ & \textbf{ASR-WER $\downarrow$} \\
\hline

LLaMa-Omni & 2.15 & 3.14($\pm$0.10) & 6.51 \\
OpenS2S\cite{wang2025opens2s} & 3.37 & 4.11($\pm$0.11) & 5.99 \\
\hline
Ours & \textbf{3.97} & \textbf{4.23($\pm$0.09)} & \textbf{5.61} \\
\hline
\end{tabular}
}
\caption{Empathetic Response Evaluation Results}
\label{tab:emotion_results}
\end{table*}

\section{EXPERIMENTS}
\subsection{Dataset Curation}
We train and evaluate the model using a combination of instruction-following and emotional dialogue speech data. To preserve general instruction-following capability in the speech modality, we use the VoiceAssistant 400k dataset ~\cite{wang2025vocalnet} and remove samples involving mathematics and symbolic reasoning, which are less suited to speech interaction in our setting. Given the scarcity of high-quality emotional dialogue speech corpora and the prohibitive cost of large-scale annotation, we construct an emotional dialogue dataset through a scalable synthesis pipeline. We first prompt GPT-4o to generate user queries across 20 domains, each paired with an emotion label (happiness, anger, fear, sadness, disgust, surprise) consistent with the query content, and then produce empathetic responses with corresponding emotion labels to ensure diversity in both text and affect. CosyVoice2 is used to synthesize emotional speech, where each label is expanded into an instructive description to leverage controllable TTS. To enhance speaker diversity, reference voices are randomly sampled from five male and five female speakers. We apply ASR-based word error rate filtering for intelligibility and a speech emotion recognizer to remove samples with mismatched affect, thereby improving alignment quality. 

To mitigate overfitting to synthetic patterns and improve generalization, we further augment with real emotional dialogues. Specifically, we perform text expansion on existing emotional TTS resources with respect to both content and emotion and synthesize corresponding empathetic responses, and we collect additional real emotional dialogue data. The resulting Emotional QA 200k dataset integrates diverse emotions, scenarios, and speakers, and is used together with VoiceAssistant 400k for training.

\subsection{Configurations}
We adopt a two-stage training strategy. Both stages use Emotional QA 200k and VoiceAssistant 400k, and AdamW optimization. In Stage 1, we fine-tune the LLM backbone with LoRA while training the downsampling adapters and the emotion prediction head. The backbone is Qwen2.5-7B-Instruct ~\cite{qwen2.5} with 28 decoder blocks and hidden size 3584, and we set the LoRA rank to 8. The emotion prediction head consists of a Transformer encoder followed by an MLP that maps the last-layer LLM hidden state to a 768-dimensional emotion vector. We use a learning rate of $5 \times 10^{-5}$ for the LoRA parameters and $2 \times 10^{-4}$ with weight decay 0.01 for the adapters and the emotion head.

In Stage 2, we train the speech decoder with six Transformer decoder blocks and a hidden size of 3584. We use a learning rate of $2 \times 10^{-4}$ and weight decay of 0.01 for the decoder.

\section{Evaluation and Results}

\subsection{General QA and Speech Quality}
We evaluate general speech understanding and instruction following on VoiceBench ~\cite{chen2024voicebench}, using the AlpacaEval, CommonEval, WildVoice, and IFEval subsets. 
Speech naturalness is assessed with UTMOS~\cite{saeki2022utmos}, a MOS prediction model. 

~\autoref{tab:performance_comparison} summarizes the results.
The proposed model achieves competitive performance across all VoiceBench subsets, indicating robust comprehension and execution from speech input, and obtains the highest UTMOS score (4.41), suggesting improved perceptual naturalness relative to baselines.

\subsection{Empathetic Response Evaluation}

We further construct a test set of 1,000 emotional queries spanning diverse scenarios and emotion categories to assess empathetic generation. At inference, the model produces both text and speech responses. For each speech response, we obtain a transcription using Whisper-large-v3 and compute WER between the ASR transcript and the model’s text output as an objective proxy for the intelligibility and semantic alignment of the speech decoder. To quantify empathy, we employ GPT-4o to judge whether the response conveys the intended emotion, attends to the user’s emotional state, and maintains semantic and emotional consistency. The judge is provided with the input query and its emotion label, the ASR transcription of the generated speech, and the predicted emotion label from a speech emotion recognizer, and assigns ratings on a 1–5 scale. We further perform a subjective Emotion MOS evaluation, in which 25 native speakers are invited to rate the generated speech in terms of naturalness and emotional expression. As shown in ~\autoref{tab:emotion_results}, our model achieves the highest Emotion GPT Score (3.97) and Speech Emotion MOS (4.23), together with the lowest WER (5.61). These results indicate that the proposed approach preserves general QA capability while improving emotional alignment and expressive quality in speech.


\subsection{Ablation Study} 

We perform an ablation to assess the effectiveness of the proposed fusion module that integrates token-level text embeddings, LLM hidden states, and emotion trajectories via AdaLN. ~\autoref{tab:ablation} reports the results in terms of Emotion GPT Score, Speech Emotion MOS, and ASR-WER. When the fusion module is removed, the decoder fails to effectively integrate contextual information from LLM hidden states with token-level text embeddings, or to modulate them through emotion trajectories via AdaLN. As a result, the model loses explicit emotion trajectory to guide expressive synthesis, leading to weaker emotional alignment and reduced naturalness.

Quantitatively, the Emotion GPT Score drops from 3.97 to 3.15, the Speech Emotion MOS drops from 4.23 to 3.85, and the ASR-WER increases from 5.61 to 6.42. These results confirm that the fusion module plays a critical role in ensuring accurate emotional expression and fluent speech generation. Subjective listening further indicates that removing text embeddings introduces pronunciation errors and prosody issues, while the lack of emotion trajectory results in speech responses that sound emotionally flat.

\begin{table}[h]
\centering
\caption{Ablation study on the fusion module.}
\label{tab:ablation}
\scalebox{0.9}{
\begin{tabular}{l|ccc}
\hline
\textbf{Model Variant} & \makecell[c]{Emotion GPT\\Score ↑} & \makecell[c]{Speech Emotion\\MOS ↑} & \makecell[c]{ASR\\WER ↓} \\
\hline
w/o Fusion Module & 3.15 & 3.85 (±0.08) & 6.42 \\
Full Model        & \textbf{3.97} & \textbf{4.23 (±0.09)} & \textbf{5.61} \\
\hline
\end{tabular}
}
\end{table}

\section{CONCLUSIONS}
\label{sec:conclusion}
In this study, we present Emotion Omni, a speech LLM that recognizes speech input and generates empathetic responses by effectively integrating semantic and emotional features. Our cost-effective emotional dialogue dataset generation pipeline reduces reliance on large-scale labeled data, making empathetic speech system development more accessible. The ability to generate high-quality empathetic speech with limited resources demonstrates the scalability and real-world potential of our approach.
While results are promising, challenges remain in handling subtle emotions, as the model sometimes prioritizes semantic content over emotional cues. Future work will focus on improving sensitivity to nuanced emotions. This research provides a foundation for building more efficient, accessible, and emotionally intelligent speech generation systems. 

\vfill\pagebreak

\bibliographystyle{IEEEbib}
\bibliography{strings,refs}

\end{document}